\documentclass[11pt,a4paper]{article}
\usepackage{authblk}
\usepackage[hyperref]{emnlp-ijcnlp-2019}
\usepackage{times}
\usepackage{latexsym}

\usepackage{url}

\usepackage{graphicx}
\usepackage{subcaption}
\usepackage{todonotes}
\usepackage{multirow}
\usepackage{booktabs}

\usepackage{amsmath}
\usepackage{amssymb}

\aclfinalcopy 

\title{Adaptive Ensembling: Unsupervised Domain Adaptation\\for Political Document Analysis}
\author[1]{{\bf Shrey Desai}}
\author[2]{{\bf Barea Sinno}}
\author[3]{{\bf Alex Rosenfeld}}
\author[3]{{\bf Junyi Jessy Li}}
\affil[1]{Department of Computer Science}
\affil[2]{Department of Government, Department of Statistics \& Data Science}
\affil[3]{Department of Linguistics}
\affil[ ]{The University of Texas at Austin}
\affil[ ]{\tt shreydesai@utexas.edu, barea.sinno@utexas.edu}
\affil[ ]{\tt alexbrosenfeld@gmail.com, jessy@austin.utexas.edu}

\date{}

\begin{document}
\maketitle
\begin{abstract}
Insightful findings in political science often require researchers to analyze documents of a certain subject or type, yet these documents are usually contained in large corpora that do not distinguish between pertinent and non-pertinent documents. In contrast, we can find corpora that label relevant documents but have limitations (e.g., from a single source or era), preventing their use for political science research. To bridge this gap, we present \textit{adaptive ensembling}, an unsupervised domain adaptation framework, equipped with a novel text classification model and time-aware training to ensure our methods work well with diachronic corpora. Experiments on an expert-annotated dataset show that our framework  outperforms strong benchmarks. Further analysis indicates that our methods are more stable, learn better representations, and extract cleaner corpora for fine-grained analysis.
\end{abstract}

\section{Introduction}

Recent progress in natural language processing and computational social science have 
pushed political science research into new frontiers. For example, scholars have studied language use in presidential elections \cite{acree2018etch}, legislative text in Congress \cite{de2018policy}, and similarities in national constitutions \cite{elkins2019evaluation}. 
However, datasets used by political scientists are mostly homogeneous in terms of subject (e.g., immigration) or document type (e.g., constitutions). Labeled corpora  with pertinent documents usually only stem from a single source; this makes it difficult to generalize  conclusions derived from them  to other sources. On the other hand, corpora spanning multiple decades and sources tend to be unlabeled. These corpora are largely untouched by political scientists; to illustrate some problems that arise with studying such data,  Table~\ref{tab:lda-intro} shows a sample of topics generated by Latent Dirichlet Allocation (LDA) \cite{blei2003latent}, a popular topic model in social science, trained on 60,000 documents sampled from the Corpus of Historical American English (COHA) \cite{davies2008corpus}. The generated topics are extremely vague and not specific to politics.

\begin{table}[]
\centering
\small
\begin{tabular}{l|l}
\toprule
Topic 1 & like, day, would, a.m., center\\ 
Topic 2 & two, samour, family, veronica, son\\ 
Topic 3 & would, hospital, also, car, hyundai\\ 
Topic 4 & said, people, one, years, think\\
Topic 5 & city, 6-4, last, wine, york\\ \bottomrule
\end{tabular}
\caption{Randomly sampled topics and top keywords derived from a 50-topic LDA model trained on a sample of COHA documents. Topic modeling results on a political subset of COHA are presented in Table~\ref{tab:ae-topics}. Additionally, topic model hyperparameters are detailed in Appendix \ref{appendix:topic}.} 
\label{tab:lda-intro}
\end{table}

This paper bridges the gap between labeled and unlabeled corpora by framing the problem as one of domain adaptation. We develop \textit{adaptive ensembling}, an unsupervised domain adaptation framework that learns from a single-source, labeled corpus (the source domain) and utilizes these representations effectively to obtain labels for a multi-source, unlabeled corpus (the target domain). Our method draws upon \textit{consistency regularization}, a popular technique that stabilizes model predictions under input or weight perturbations \cite{athiwaratkun2018there}. At the framework-level, we introduce an adaptive, feature-specific approach to optimization; at the model-level, we develop a novel text classification model that works well with our framework. To better handle the diachronic nature of our corpora, we also incorporate time-aware training and representations.

Our experiments use the New York Times Annotated Corpus (NYT) \cite{sandhaus2008new} as our source domain corpus and COHA as our target domain corpus. Concretely, we construct two classification tasks: a \textit{binary task} to determine whether a document is political or non-political; and a \textit{multi-label task} to categorize a document under three major areas of political science in the US: \textit{American Government}, \textit{Political Economy}, and \textit{International Relations} \cite{goodin2009oxford}. We subsequently introduce an expert-labeled test set from COHA to evaluate our methods.

Our framework, equipped with our best model, significantly outperforms existing domain adaptation algorithms on our tasks. In particular, adaptive ensembling achieves gains of  11.4 and 10.1 macro-averaged F1 on the binary and multi-label tasks, respectively. Qualitatively, adaptive ensembling conditions the optimization process, learns smoother latent representations, and yields precise but diverse topics as demonstrated by LDA on an extracted political subcorpus of COHA.  We release our code and datasets at \url{http://github.com/shreydesai/adaptive-ensembling}.

\section{Motivation from Political Science} \label{sec:task}
Quantitative studies of American public opinion over time have mostly been restricted to surveys such as the American National Election Survey  \cite{baldassarri2008partisans,campbell1980american}. 
However, surveys often do not pose well-formed questions, reflect true voter opinion, or capture mass public opinion  \cite{zaller1992nature,bishop2004illusion}. Therefore, researchers often seek to compare survey findings with those of mass media 
as the relationship between public opinion and the media has been widely established \cite{ baum2008relationships, mccombs2018setting}.  Press media, one form of mass media, manifests itself in large, diachronic collections of newspaper articles; such corpora provide a promising avenue for studying public opinion and testing theories, provided scholars can be confident that the measures they obtain over time are substantively invariant \cite{davidov2014measurement}. However, as alluded to earlier,  such diachronic corpora are often unlabeled; 
political scientists cannot draw conclusions from these corpora in their raw form as they are unable to distinguish between political and non-political articles. We frame this problem as an exchange between two domains: a source, labeled corpus with modern articles (NYT) and a target, unlabeled corpus with decades of articles originating from a multitude of news sources (COHA). Using domain adaptation methods, we can extract a political subcorpus from COHA that would be amenable for the study of public opinion research over time.

\section{Unsupervised Domain Adaptation}

In this section, we detail the core concepts behind our unsupervised domain adaptation framework. We describe the problem setup (\S\ref{sec:setup}), an overview of self-ensembling and consistency regularization (\S\ref{sec:cons-reg}-\S\ref{sec:fixed-ens}), and our novel contributions to this framework (\S\ref{sec:adaptive-ens}-\S\ref{temp-bias}). 

\subsection{Problem Setup} \label{sec:setup}

Let $\mathcal{X}$ and $\mathcal{Y}$ denote the input and output spaces, respectively. 
We have access to labeled samples $\{x^{(i)}_L, y^{(i)}_L\}^N_{i=1}$ from a source domain $\mathcal{D}_S$ and unlabeled samples $\{x^{(i)}_U\}^M_{i=1}$ from a target domain $\mathcal{D}_T$. The goal of unsupervised domain adaptation is to learn a function $f: \mathcal{X} \rightarrow \mathcal{Y}$ that maximizes the likelihood of the target domain samples by only leveraging supervision from the source domain samples. We also assume the existence of a small amount of labeled target domain samples in order to create a development set, following existing work in unsupervised domain adaptation
\cite{glorot2011domain,chen2012marginalized,french2017self,zhang2017aspect}.

\subsection{Self-Ensembling} \label{sec:cons-reg}

Our unsupervised domain adaptation framework builds on top of \textit{self-ensembling} \cite{laine2016temporal}, a semi-supervised learning algorithm based on \textit{consistency regularization}, whereby models are trained to be robust against injected noise \cite{athiwaratkun2018there}.

\begin{figure}[t]
    \centering
    \includegraphics[scale=0.35]{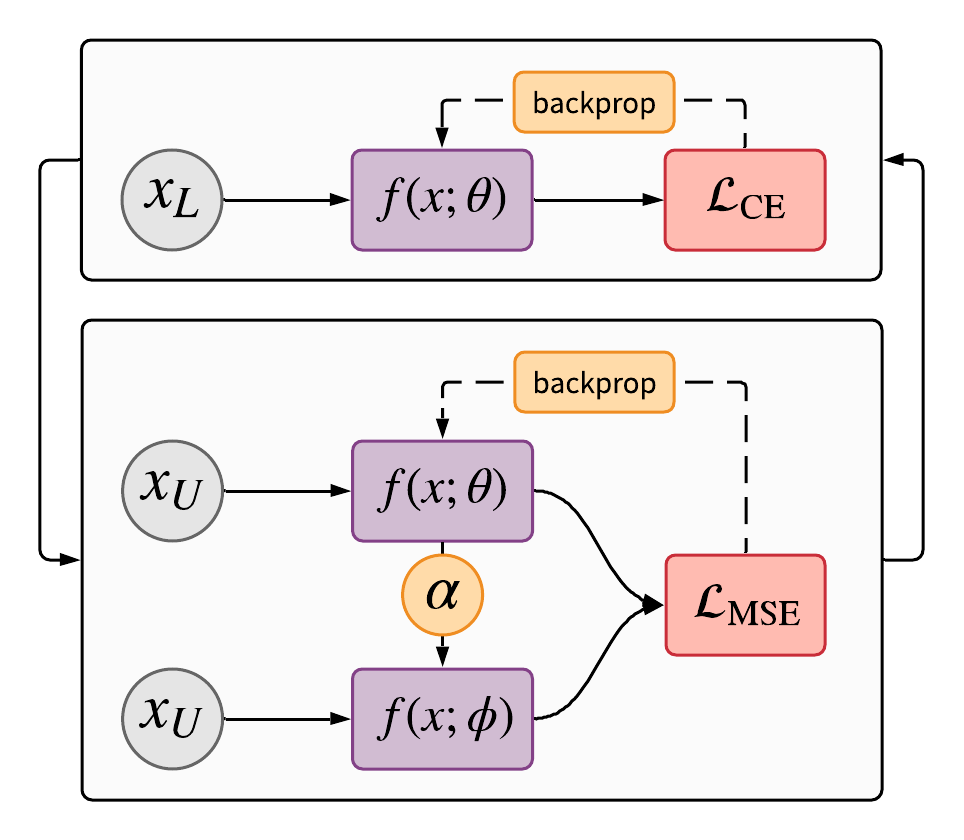}
    \caption{Visualization of the self-ensembling training procedure. Descriptions of individual components are detailed in \S\ref{sec:cons-reg}-\S\ref{sec:fixed-ens}.}
    \label{fig:fw-img}
\end{figure}

Self-ensembling is an interplay between two neural networks: a student network $f(x;\theta)$ and a teacher network $f(x;\phi)$. The inputs to both networks are perturbed separately, and the objective is to measure the consistency of the student network's predictions against the teacher's. Both networks share the same base model architecture and initial parameter values, but follow different training paradigms \cite{laine2016temporal}. In particular, the student network is updated via backpropagation, then the teacher network is updated with an exponential average of the student network's parameters \cite{tarvainen2017mean}. The networks are trained in an alternating fashion until they converge. During test time, the teacher network is used to infer the labels for target domain samples. Figure \ref{fig:fw-img} visualizes the overall training procedure. Further intuition behind self-ensembling is available in Appendix \ref{sec:se-info}.

Next, we discuss the training process for the student network (\S\ref{sec:student-training}), the original fixed ensembling method in \citet{tarvainen2017mean} (\S\ref{sec:fixed-ens}), and our proposed adaptive ensembling method (\S\ref{sec:adaptive-ens}).

\subsection{Student Training} \label{sec:student-training}
The student network uses labeled samples from the source domain and unlabeled samples from the target domain to learn domain-invariant features. This is realized by using multiple loss functions, each with its own objective. The \textbf{supervised loss} is simply the cross-entropy loss of the student network outputs given source domain samples:
\begin{equation*}
    \mathcal{L}_\textnormal{CE}(\theta) = \sum_{(x, y) \in \mathcal{D}_S} \log p(y|x')
    \label{ce_loss}
\end{equation*}
However, the supervised loss alone 
prevents the student network from learning anything useful about the target domain. To address this, \citet{laine2016temporal} introduce an \textbf{unsupervised loss} to ensure that the student and teacher networks have similar predictions for target domain samples. 
\citet{french2017self} only enforce the consistency constraint for target domain samples, but we propose using both source \textit{and} target domain samples with separately perturbed inputs $x'$ and $x''$; this provides a balanced source of supervision to train our adaptive constants, discussed in \S\ref{sec:adaptive-ens}:
\begin{equation*}
    \mathcal{L}_\textnormal{MSE}(\theta, \phi) = \sum_{x \in \mathcal{D}_S \cup \mathcal{D}_T} ||f(x';\theta) - f(x'';\phi)||^2
\end{equation*}
\vspace{-1em}

The overall objective is a combination of the two loss functions:
\begin{equation*}
    \mathcal{L}(\theta, \phi) = \mathcal{L}_\textnormal{CE} + \mathcal{L}_\textnormal{MSE}
\end{equation*}

\subsection{Fixed Ensembling} \label{sec:fixed-ens}

The teacher network's parameters form an ensemble of the student network's parameters over the course of training:
\begin{equation*}
    \phi^{(t+1)} \leftarrow \alpha \phi^{(t)} + (1-\alpha) \theta^{(t)}
\end{equation*}

\noindent where $\alpha$ is a smoothing factor that controls the magnitude of the parameter updates. Since the labels for the target domain samples are inherently unknown, ensembling parameters in the presence of noise helps the teacher network's predictions converge to the \textit{true} label \cite{tarvainen2017mean}.

\paragraph{Limitations} Empirically, we find that the highly unstable loss surface presented by textual datasets causes large instabilities in the optimization process. One of the key insights of this paper is that these instabilities are due to the dynamics of the unsupervised loss. Because the unsupervised loss effectively regularizes the source domain representations to work well in the target domain \cite{laine2016temporal}, performance degrades rapidly if this loss fails to converge. This is a strong indicator that self-ensembling fails to learn useful, shared representations for knowledge transfer between textual domains. Qualitative evidence of the unsupervised loss' instability is shown in Figure~\ref{fig:unsup_loss} and further discussed in \S\ref{sec:analysis}.

\subsection{Adaptive Ensembling} \label{sec:adaptive-ens}

We hypothesize that smoothing with a fixed hyperparameter $\alpha$ is responsible for said instabilities. For any given weight matrix (or bias vector), each hidden unit can be conceptualized as controlling one highly specific feature or attribute \cite{bau2018identifying}. 
These units may need to be updated with varying degrees throughout the course of training; therefore, smoothing each unit with a fixed constant severely overlooks dynamics at the parameter-level. We propose modifying fixed ensembling by introducing trainable smoothing constants for each unit---hereafter termed \textit{adaptive constants}---as opposed to using a fixed smoothing constant:
\begin{equation*}
    \phi^{(t+1)} \leftarrow \mathbf{C}^{(t)} \odot \phi^{(t)} + (\mathbf{1} - \mathbf{C}^{(t)}) \odot \theta^{(t)}
\end{equation*}
\noindent where a matrix of adaptive constants $\mathbf{C}$ is applied element-wise to $\phi$ and $\theta$ at each step.

\paragraph{Example} Assume we are training an arbitrary weight matrix $\mathbf{W} \in \mathbb{R}^{m \times n}$ in the $k$th layer of a fixed network architecture. Both the student and teacher network have their own copy of $\mathbf{W}$, denoted as $\mathbf{W}_\textnormal{STU}$ and $\mathbf{W}_\textnormal{TEA}$, respectively. To ensure each parameter $\mathbf{W}_{ij}$ has a corresponding adaptive constant $\alpha_{ij}$, $\mathbf{C}$ shares the same dimensionality as $\mathbf{W}_\textnormal{STU}$ and $\mathbf{W}_\textnormal{TEA}$. The previous equation can then be written as:
\begin{equation*}
    \mathbf{W}^{(t+1)}_\textnormal{TEA} \leftarrow \mathbf{C}^{(t)} \odot \mathbf{W}^{(t)}_\textnormal{TEA} + (\mathbf{1} - \mathbf{C}^{(t)}) \odot \mathbf{W}^{(t)}_\textnormal{STU}
\end{equation*}

\paragraph{Supervision} Because the adaptive constants are designed to stabilize training, it is a natural fit to train them using the unsupervised loss:
\begin{equation*}
    \mathbf{C}^{(t+1)} \leftarrow \mathbf{C}^{(t)} - \epsilon\nabla_\mathbf{C} \mathcal{L}_\textnormal{MSE}
\end{equation*}
This forms a crucial difference between self-ensembling and adaptive ensembling: in the former method, the teacher network has no say in how its parameters are modified. Adaptive ensembling equips the teacher network with fine-grained control over gradient updates, making it far easier to align activations under a noisy setting.

\subsection{Temporal Curriculum} \label{temp-bias}
\begin{figure}
    \centering
    \includegraphics[scale=0.5]{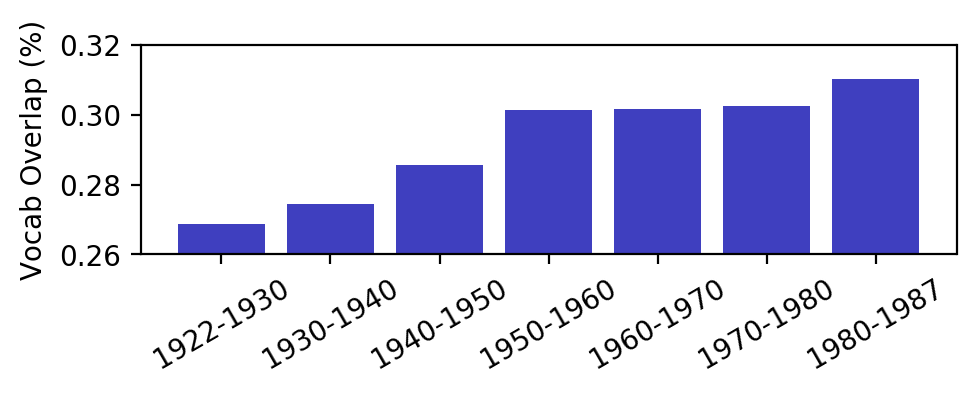}
    \caption{Vocabulary overlap between COHA and NYT, by decade. We collect COHA documents in each decade, create a decade vocabulary, and calculate the percentage overlap between each decade's vocabulary and the overall NYT vocabulary.}
    \label{fig:vocab}
\end{figure}

Diachronic datasets important in political science can be difficult to adapt to given the minimal vocabulary overlap between the source and target domain documents. 
Source and target articles mention named entities and events that, for the most part, do not appear across both datasets. To ease the difficulty of domain adaptation, we exploit the temporal information in our datasets to introduce a curriculum \cite{bengio2009curriculum}.

In particular, each article comes with metadata that includes the year in which the article was published. Figure \ref{fig:vocab} shows that COHA articles written closer to the time of NYT articles have a larger vocabulary overlap than those written in the distant past. Intuitively, it is easier to learn features from target domain samples that are more like the source domain samples. Hence, we sort the target domain mini-batches by year; the learning task becomes progressively harder as opposed to confusing the models during the early stages of training.

\section{Model}

In this section, we introduce a new convolutional neural network (CNN) as the plug-in model for our unsupervised domain adaptation framework. We motivate the use of CNNs (\S\ref{sec:model-motiv}), formalize the model input (\S\ref{sec:model-input}), and introduce several novel components for our task (\S\ref{sec:model-arch}).

\subsection{Motivation} \label{sec:model-motiv}

CNNs have emerged as strong baselines for text classification in NLP \cite{kim2014convolutional}. CNNs are desirable candidates for our framework as they exhibit a high degree of parameter sharing, significantly reducing the number of parameters to train. In addition, they can be designed to solely optimize the log-likelihood of the training data. Experimentally, we find that models that optimize other distributions (e.g., attention distributions in Transformers \cite{vaswani2017attention} or Hierarchical Attention Networks \cite{yang2016hierarchical}) do not work well with this framework.

\subsection{Model Input} \label{sec:model-input}

Given a discrete input $x = [w_1, \cdots, w_n]$ and vocabulary $V$, an embedding matrix $\mathbf{E} \in \mathbb{R}^{|V| \times d}$ replaces each word $w_i$ with its respective $d$-dimensional embedding. The resulting embeddings are stacked row-wise to obtain an input matrix
$\mathbf{X} \in \mathbb{R}^{n \times d}$. Following the notion of input perturbation used in consistency regularization algorithms \cite{athiwaratkun2018there}, we design several methods to inject noise into the input layer. Each input is perturbed with additive, isotropic Gaussian noise: $\tilde{\mathbf{X}} = \mathbf{X} + \mathcal{N}(0, \mathbf{I})$. Then, we apply dropout on the perturbed inputs to eliminate dependencies on any one word: $\mathbf{X'} = \tilde{\mathbf{X}} \odot \mathbf{M}$ where $\mathbf{M} \in \mathbb{R}^{n \times d}$ is a Bernoulli mask applied element-wise to the input matrix.

\subsection{Model Architecture} \label{sec:model-arch}

\paragraph{Background: 1D Convolutions} CNNs for text classification generally use 2D convolutions over the input matrix \cite{kim2014convolutional}, but architectures using 1D convolutions have also been explored in other contexts, e.g., sequence modeling \cite{bai2018empirical}, machine translation \cite{kalchbrenner2016neural}, and text generation \cite{yang2017improved}. Our model draws upon the latter approach for political document classification. CNNs utilizing 1D convolutions are typically autoregressive in nature; that is, each output $y_t$ only depends on the inputs $x_{<t}$ to avoid information leakage into the future. Two approaches have been proposed to achieve this: history-padding \cite{bai2018empirical,bai2018trellis} and masked convolutions \cite{kalchbrenner2016neural}. Further, each successive convolution uses an exponentially increasing dilation factor, reducing the depth of the network significantly. Below, we elaborate on the components of our model:

\paragraph{Sequence Squeezing} Given a model with $\ell$ layers, previous approaches \cite{bai2018empirical,bai2018trellis} history-pad the input with $\sum_{i=1}^{\ell}d^{(i-1)}(f-1)$ zeros to obtain an output of length $n$, where $d$ is the dilation factor and $f$ is the filter size. However, we propose history-padding the input with $(\sum_{i=1}^{\ell}d^{(i-1)}(f-1))-n+1$ zeros to ensure the convolutions compress the sequence down to \textit{one} output unit. Formally, this produces an output feature map of dimension $B \times C \times 1$ where $B$ is the batch size and $C$ is the number of channels; one can use a simple $\texttt{squeeze()}$ operation to obtain the compact feature matrix $B \times C$. Though this is a subtle difference, our approach yields much richer representations for classification.

\begin{figure}
    \centering
    \includegraphics[scale=0.4]{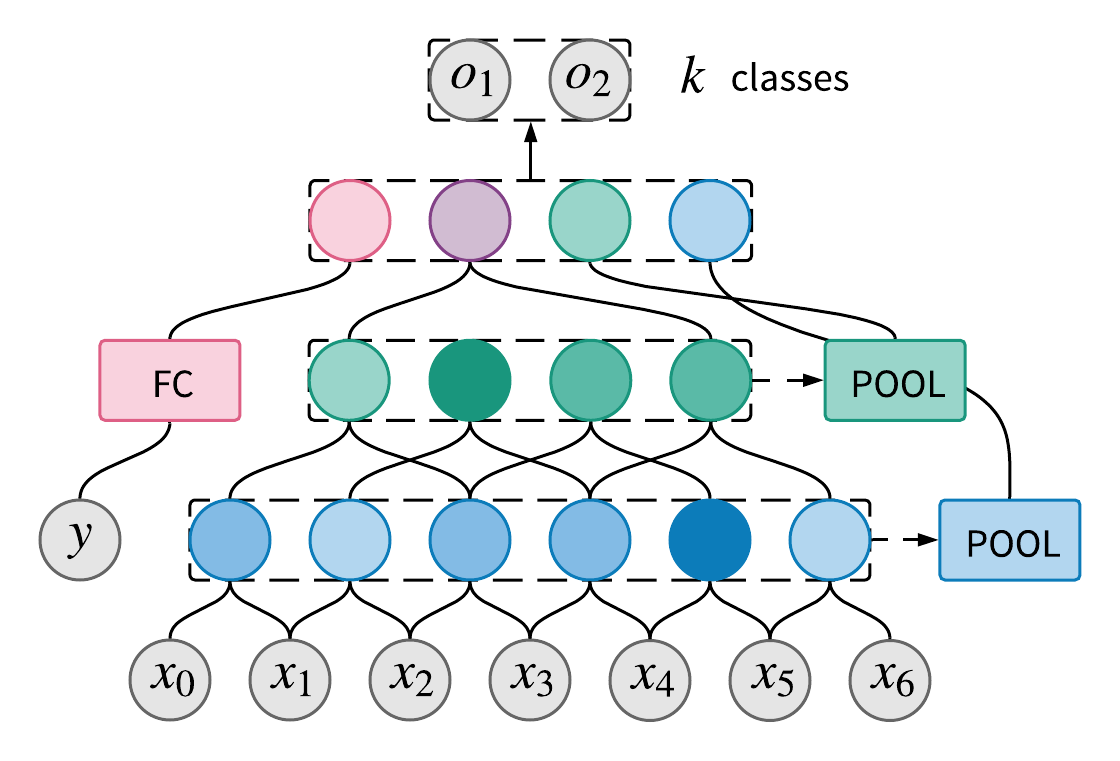}
    \caption{Example three-layer CNN architecture with sequence squeezing, state connections, and time embeddings. Detailed information about individual components is available in \S\ref{sec:model-arch}. FC represents a fully-connected layer; POOL represents a pooling layer. Best viewed in color.}
    \label{fig:model}
\end{figure}

\paragraph{State Connections} In each layer $\ell_i$, a kernel $\mathbf{W}_i$ convolves across an intermediate sequence, inducing a feature map $\mathbf{A}_i$. Because the input is presented as a sequence, the application of $\mathbf{W}_i$ along a one-dimensional axis encourages $\mathbf{A}_i$ to encode temporal features, similar to how the hidden state is formed by applying shared weights across a sequence in recurrent architectures. Further, because the receptive field grows exponentially, the convolutions build hierarchical representations of the input, implying $\mathbf{A}_{i+1}$ builds a more abstract representation of the input than $\mathbf{A}_i$. We exploit this stateful information by pooling each activation map $\mathbf{A}_i$ into a vector and concatenating them row-wise to create a \textit{state matrix}:
\begin{equation*}
    \mathbf{S} = \textnormal{Concat}(\textnormal{Pool}(\mathbf{A}_1), \cdots, \textnormal{Pool}(\mathbf{A}_{\ell-1}))
\end{equation*}
To the best of our knowledge, our paper is the first to explicitly use the temporal state embedded in causal 1D convolution activations as representations for an end task.

\paragraph{Time Embedding}

To make our model time aware, we learn representations for the years of the documents (available as metadata in COHA). Such time representations allow the model to reason about content as it appears in different decades. Given a year $y$ (e.g. 1954), we normalize it to the closed unit interval $[0,1]$ and linearly transform it into a low-dimensional embedding $\mathbf{e}$:
\begin{equation*}
    \mathbf{e} = \mathbf{W}_e\bigg[\frac{y - \textnormal{max}_y}{\textnormal{max}_y-\textnormal{min}_y}\bigg]+b_e
\end{equation*}

\noindent where $\textnormal{max}_y$ and $\textnormal{min}_y$ represent the maximum and minimum observed years in the training dataset, respectively.

\paragraph{Overall Architecture} We concatenate the various components of our model $[\mathbf{X}';\mathbf{S};\mathbf{e}]$ to create a collective representation for classification. We use a 1D convolution ($f=1$ and $d=1$) to project this representation to $k$ classes:
\begin{equation*}
    \mathbf{y} = \textnormal{Conv1D}([\mathbf{X}';\mathbf{S};\mathbf{e}]; \mathbf{W}_k)
\end{equation*}
We did not observe any performance advantages from using a fully-connected layer to perform the projection, so we opt to use a fully-convolutional architecture to minimize the number of parameters \cite{long2015fully}. Finally, we apply softmax to the output vector $\mathbf{y} \in \mathbb{R}^k$ to obtain a valid probability distribution over the classes. An example of our model architecture is depicted in Figure~\ref{fig:model}.

\section{Datasets}\label{sec:data}

We present a dataset for identifying political documents with manual annotation from political science graduate students.
The dataset is constructed for \textit{binary} and \textit{multi-label} tasks: (1) identifying whether a document is political (i.e. containing notable political content) and (2) if so, the area(s) among three major political science subfields in the US: {\em American Government}, {\em Political Economy}, and {\em International Relations}~\cite{goodin2009oxford}.

\paragraph{Source} We use NYT as the source dataset as it contains fine-grained descriptors of article content. We sample 4,800 articles with the descriptor \textsc{US Politics \& Government}. To obtain non-political articles, we sample 4,800 documents whose descriptors do not overlap with an exhaustive list of political descriptors identified by a political science graduate student. For our multi-label task, the annotator grouped descriptors in NYT that belong to each area label we consider\footnote{These descriptors are available in Appendix \ref{sec:nyt-data}.}.

\paragraph{Target} Our target data are historical documents from COHA, which contains a large collection of news articles since the 1800s. To ensure our dataset is useful for diachronic analysis (e.g., public opinion over time), we sample only from news sources that consistently appear across the decades. Further, we ensure there are at least 8,000 total documents in each decade group; this narrows down our time span to 1922--1986. From this subset, we sample $\sim$250 documents from each decade for annotation. Two political science graduate students each annotated a subset of the data.

To train our unsupervised domain adaptation framework, we use 9,600 unlabeled target examples (same number as NYT). The expert-annotated dataset is divided into three subsets: a training set of 984 documents ({\em only} for training the In-Domain classifier discussed in \S\ref{sec:results-framework}), development set of 246 documents, and test set of 350 documents (50 per decade)\footnote{The news sources used and label distributions for the expert-annotated dataset are available in Appendix \ref{sec:expa-data}.}.

\section{Experiments}
\subsection{Settings}\label{sec:settings}

Our CNN has $8$ layers, each with $256$ channels, $f=3$, $d=2^i$ (for the $i$th layer), and $\textnormal{ReLU}$ activation. We enforce a maximum sequence length of $200$ and minimum word count from $[1,2,3]$ to build the vocabulary. The embedding matrix uses $300$-D GloVe embeddings \cite{pennington2014glove} with a dropout rate of $0.5$ \cite{srivastava2014dropout}. We history-pad our input with a zero vector, the state connections are obtained using average pooling, and the time embedding has a dimensionality of $10$. The model is optimized with Adam \cite{kingma2014adam}, learning rate from $[10^{-4}, 5\cdot 10^{-5}, 10^{-5}]$, and mini-batch size from $[16,32]$. Hyperparameters were discovered using a grid search on the held-out development set.

\begin{table}[]
\centering
\small
\begin{tabular}{l|cc|ccc}
\toprule
 & \multicolumn{2}{c|}{Binary Task} & \multicolumn{3}{c}{Multi-Label Task} \\ \midrule
Method & Mi-F & Ma-F & Ma-P & Ma-R & Ma-F \\ \midrule
Source Only & 55.7 & 46.2 & 28.8 & 70.0 & 39.6 \\ 
mSDA & 57.4 & 49.7 & 41.0 & 63.7 & 48.1 \\ 
DANN & 68.2 & 65.8 & \textbf{50.8} & 36.3 & 42.2 \\
SE & 64.0 & 59.5 & 42.7 & 64.1 & 51.0 \\ 
~~+ curriculum & 66.4 & 62.3 & 44.4 & 71.7 & 54.5 \\
AE (ours) & 75.1 & 74.5 & 46.1 & 75.3 & 57.2 \\
~~+ curriculum & \textbf{77.4} & \textbf{77.1} & 48.2 & \textbf{83.5} & \textbf{61.1} \\ \midrule
In-Domain & 81.7 & 81.6 & 86.5 & 83.5 & 84.8 \\ \bottomrule
\end{tabular}
\caption{Framework results for the binary label task (left) and multi-label task (right). For the binary task, we show micro- and macro-averaged F1 scores. For the multi-label task, we show macro-averaged precision, recall, and F1 scores.}
\label{tab:fw-results}
\end{table}

\subsection{Framework Results}\label{sec:results-framework}

Using our best model, we benchmark our unsupervised domain adaptation framework against established methods: (1) \textbf{Marginalized Stacked Denoising Autoencoders (mSDA):} Denoising autoencoders that marginalize out noise, enabling learning on \textit{infinitely many} corrupted training samples \cite{chen2012marginalized}. (2) \noindent \textbf{Self-Ensembling (SE):} A consistency regularization framework that stabilizes student and teacher network predictions under injected noise (discussed in detail in \S\ref{sec:cons-reg}-\S\ref{sec:fixed-ens}) \cite{laine2016temporal,tarvainen2017mean,french2017self}. (3) \textbf{Domain-Adversarial Neural Network (DANN):} Multi-component framework that learns domain-invariant representations through adversarial training \cite{ganin2016domain}. We also benchmark against \textbf{Source Only} (classifier trained on the source domain only) and \textbf{In-Domain} (classifier trained on the target domain only) to establish lower and upper performance bounds, respectively \cite{zhang2017aspect}.

Framework results are presented in Table \ref{tab:fw-results}. Our method achieves the highest F1 scores for both tasks. The temporal curriculum further improves our results by a large margin, validating its effectiveness for domain adaptation on diachronic corpora. Although DANN achieves higher precision on the multi-label task, its recall largely suffers.

\begin{table}[]
\centering
\small
\begin{tabular}{l|cc|ccc}
\toprule
 & \multicolumn{2}{c|}{Binary Task} & \multicolumn{3}{c}{Multi-Label Task} \\ \midrule
Model & Mi-F & Ma-F & Ma-P & Ma-R & Ma-F \\ \midrule
LR & 63.7 & 60.9 & 53.3 & 67.3 & 31.9 \\ 
BiLSTM & 64.8 & 63.1 & 36.2 & 65.0 & 46.3 \\ 
CNN (2D) & 73.1 & 72.1 & \textbf{49.0} & 73.8 & 58.9 \\ 
CNN (ours) & 75.4 & 75.3 & 36.9 & \textbf{91.4} & 52.5 \\ 
+ seq squeeze & 75.1 & 74.6 & 45.8 & 79.6 & 58.2 \\ 
~~+ state conn & \textbf{80.2} & 76.3 & 45.3 & 85.5 & 59.2 \\ 
~~~~+ time emb & 77.4 & \textbf{77.1} & 48.2 & 83.5 & \textbf{61.1} \\ \bottomrule
\end{tabular}
\caption{Model results with adaptive ensembling for the binary label task (left) and multi-label task (right). For the binary task, we show micro- and macro-averaged F1 scores. For the multi-label task, we show macro-averaged precision, recall, and F1 scores.}
\label{tab:model-results}
\end{table}

\subsection{Model Results}\label{sec:results-model}

Next, we ablate the various components of our model and evaluate several other strong text classification baselines under our framework: (1) \textbf{Logistic Regression (LR):} We average the word embeddings of each token in the sequence, then use these to train a logistic regression classifier. (2) \textbf{Bidirectional LSTM (BiLSTM):} A bidirectional LSTM obtains forwards and backwards input representations $[\overrightarrow{\mathbf{h}_t};\overleftarrow{\mathbf{h}_t}]$ \cite{hochreiter1997long}; they are concatenated and passed through a fully-connected layer. (3) \textbf{CNN (2D):} A CNN using 2D kernels $3 \times 300$, $4 \times 300$, and $5 \times 300$ obtains representations \cite{kim2014convolutional}. They are max-pooled, concatenated row-wise, and passed through a fully-connected layer.

Model ablations and results are presented in Table \ref{tab:model-results}. Our full model achieves the highest F1 scores on both the binary and multi-label tasks, and each  component consistently contributes to the overall F1 score. The 2D CNN also has decent F1 scores, showing that our framework works with standard CNN models. Further, the time embedding significantly improves both F1 scores, indicating the model effectively utilizes the unique temporal information present in our corpora.

\section{Analysis} \label{sec:analysis}

In this section, we pose and qualitatively answer numerous probing questions to further understand the strong performance of adaptive ensembling. We analyze several characteristics of the overall framework (\S\ref{sec:fw}), then qualitatively inspect its performance on our datasets (\S\ref{sec:ds}).

\subsection{Framework} \label{sec:fw}

\begin{figure}
    \centering
    \includegraphics[scale=0.5]{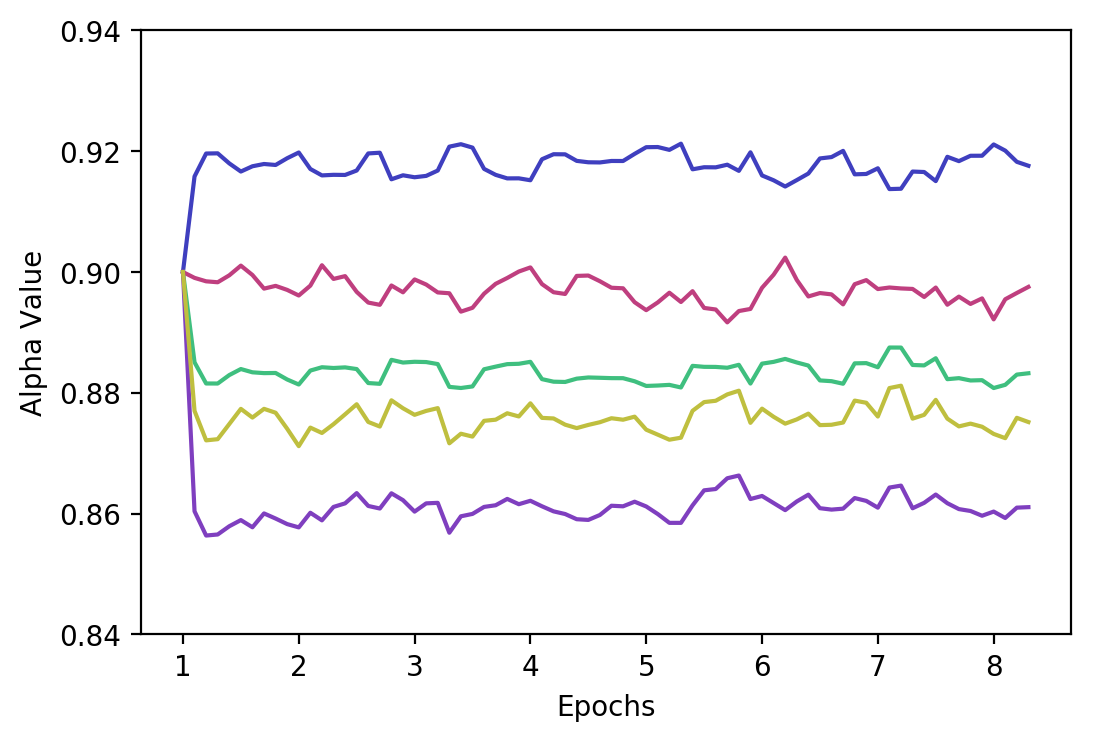}
    \caption{Trajectories of five randomly sampled adaptive constants from our CNN over the course of training. Best viewed in color.}
    \label{fig:trajectories}
\end{figure}

\paragraph{Are the adaptive constants different across hidden units?} We randomly sample five adaptive constants and track their value trajectories over the course of training. Figure \ref{fig:trajectories} shows all of them sharply converge to and bounce around the same general neighboorhood. This is strong evidence that we cannot use a fixed hyperparameter $\alpha$ to smooth each parameter, rather we need per-parameter smoothing constants to account for the functionality and behavior of each unit.

\begin{figure}
    \centering
    \includegraphics[scale=0.50]{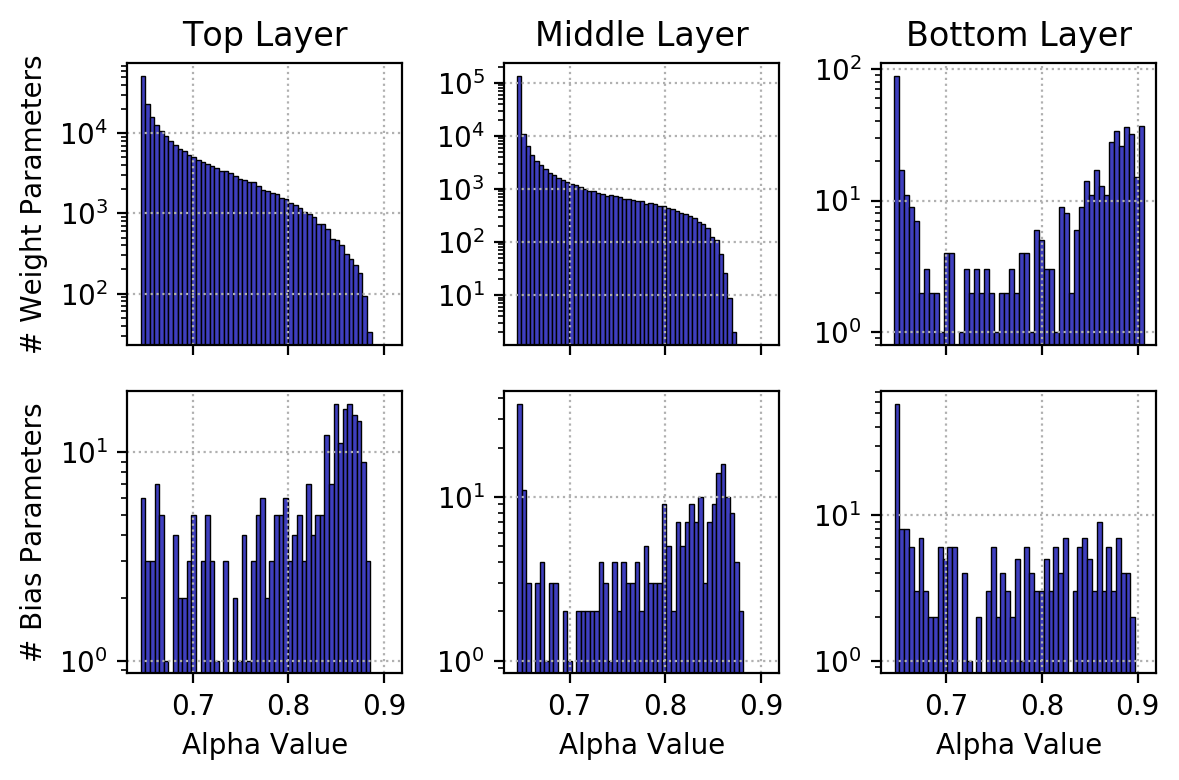}
    \caption{Distribution of teacher network adaptive constants for a top, middle, and bottom layer. We display adaptive constants for both weight (top) and bias (bottom) parameters. The $x$-axis is shared for both the weight and bias distributions.}
    \label{fig:layer_distrib}
\end{figure}

\paragraph{How do the adaptive  constants change by layer?}
Figure ~\ref{fig:layer_distrib} shows the distribution of weight and bias parameters of adaptive constants for a top, middle, and bottom layer of our CNN. For the weight parameters, the teacher relies heavily on the student ($\alpha$ is skewed towards smaller smoothing rates) in the top layer, but gradually reduces its dependence by learning target domain features in the lower layers ($\alpha$ is skewed towards larger smoothing rates). For the bias parameters, the teacher prominently shifts the student features to work for the target domain in the top layer, but reduces its dependence on the student in the lower layers. This shows why using a fixed hyperparameter $\alpha$ does not account for layer-wise dynamics, i.e. each layer requires a specific distribution of $\alpha$ values to achieve strong performance.

\begin{figure}[t]
\centering
\begin{subfigure}[t]{\columnwidth}
\centering
    \includegraphics[width=6.5cm]{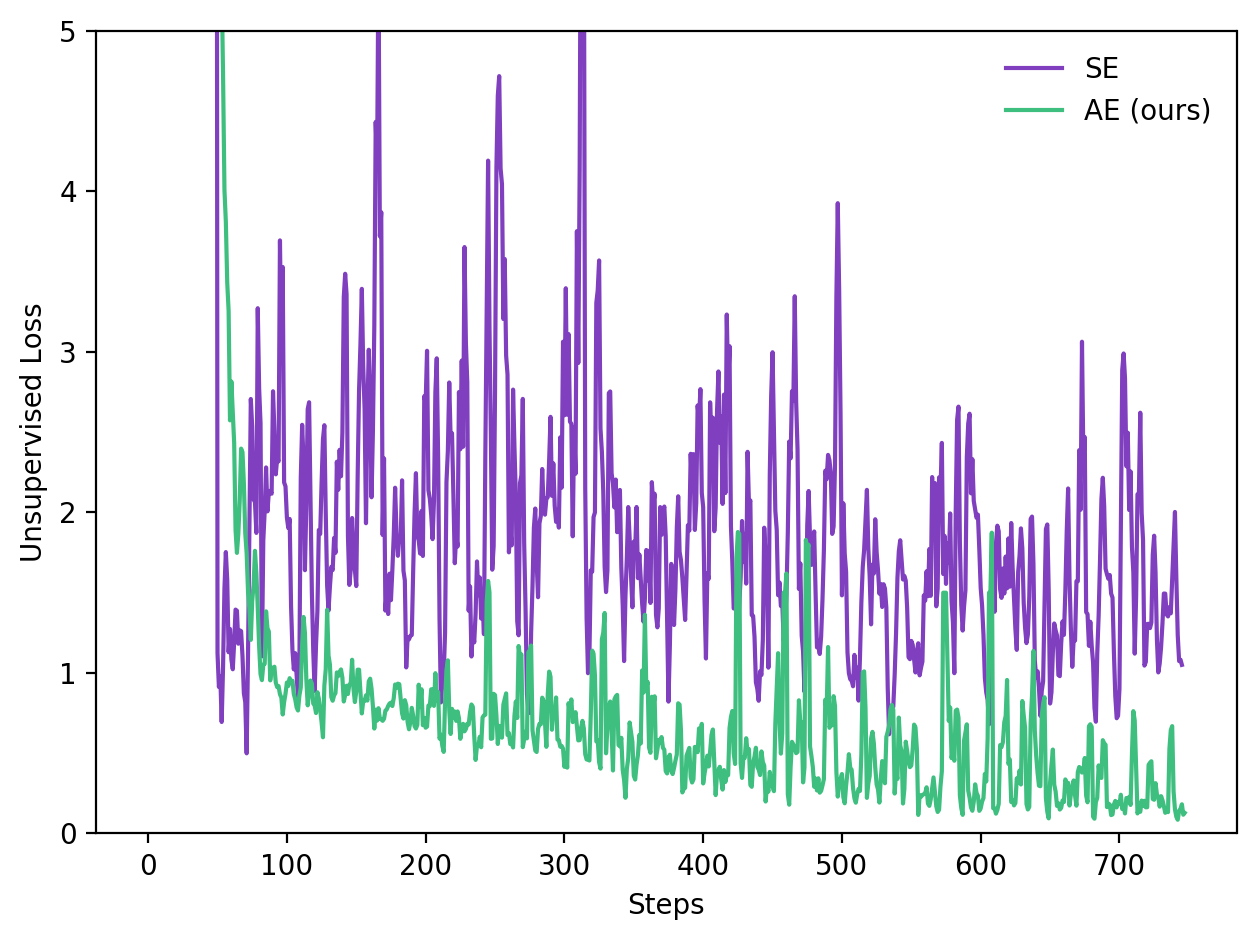}
    \caption{Loss curves for self-ensembling (SE) and adaptive ensembling (AE) over the course of training.}
    \label{fig:unsup_loss}
\end{subfigure}

\begin{subfigure}[t]{\columnwidth}
\centering
    \includegraphics[width=5.8cm]{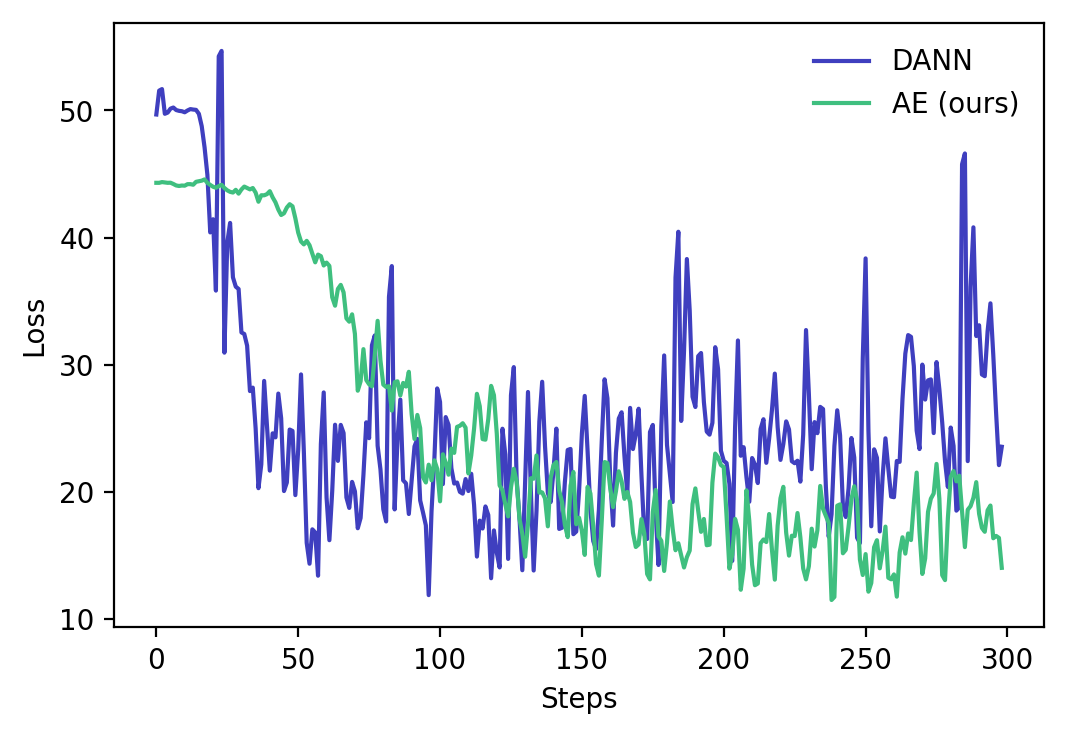}
    \caption{Loss curves for domain-adversarial neural networks (DANN) and adaptive ensembling (AE). The adversarial loss weight and unsupervised loss weight are annealed for both methods. For fair comparison, we employ the adversarial weight annealing schedule outlined in \citet{ganin2016domain}.}
    \label{fig:comparison}
\end{subfigure}
\end{figure}

\begin{figure}[t]
    \centering
    \includegraphics[scale=0.5]{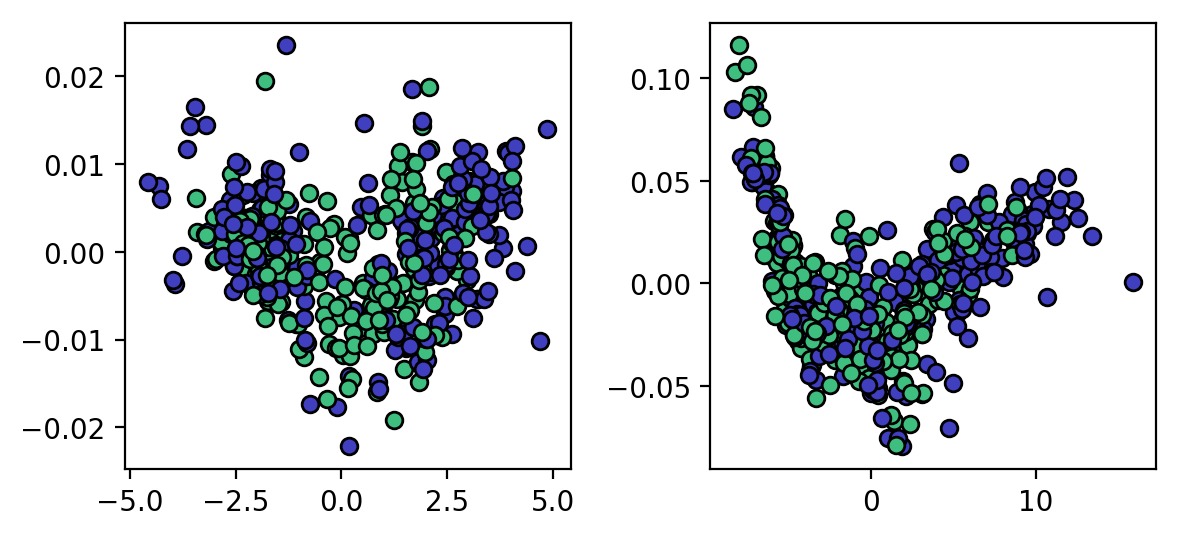}
    \caption{PCA performed on the latent representations of the teacher network in self-ensembling (left) and adaptive ensembling (right). We show representations for both source domain samples (green) and target domain samples (blue). Best viewed in color.}
    \label{fig:latent_space}
\end{figure}

\paragraph{Do adaptive constants benefit training and latent representations?}
Figure~\ref{fig:unsup_loss} depicts the unsupervised loss trajectories for self-ensembling (SE) and adaptive ensembling (AE). Compared to SE, the adaptive constants significantly stabilize the unsupervised loss. Next, Figure~\ref{fig:comparison} shows the general training curves for AE and domain-adversarial neural networks (DANN). The DANN loss oscillates uncontrollably as the adversarial weight increases, but increasing the unsupervised loss weight for AE does not result in as much instability. We also compare the latent representations learned by SE and AE in Figure~\ref{fig:latent_space}. While SE shows evidence of feature alignment, AE learns a much smoother manifold where source and target domain representations are intertwined.

\subsection{Datasets} \label{sec:ds}

\paragraph{Does adaptive ensembling yield better topics?}
In Table~\ref{tab:lda-intro}, we showed that applying LDA directly on COHA yields noisy, unrecognizable topics. Here, we use the \textsc{Source Only} model and the adaptive ensembling framework to obtain labels for the unlabeled pool of COHA documents. We extract the political documents, run a topic model on the political subcorpus, and randomly sample topics. The \textsc{Source Only} results are shown in Table \ref{tab:nyt-topics} and the adaptive ensembling results are shown in Table \ref{tab:ae-topics}. The \textsc{Source Only} model has poor recall, as most of the topics are extracted are vague and not inherently political in nature. In contrast, our framework is able to extract a wide range of clean, identifiable political topics. For example, the first topic reflects documents related to the Vietnam conflict while the third topic reflects documents related to important court proceedings.

\begin{table}[t]
\centering
\small
\begin{tabular}{l|l}
\toprule
Topic 1 & dr, women, week, medical, doctors\\
Topic 2 & city, police, street, car, avenue\\ 
Topic 3 & trial, years, police, prison, court\\ 
Topic 4 & union, strike, workers, lewis, service\\ 
Topic 5 & like, man, years, little, week\\ \bottomrule
\end{tabular}
\caption{Randomly sampled topics and top keywords derived from a 50-topic LDA model trained on 28K COHA articles identified as political using the \textsc{Source Only} model.} 
\label{tab:nyt-topics}
\end{table}

\begin{table}[t]
\centering
\small
\begin{tabular}{l|l}
\toprule
Topic 1 & vietnam, hanoi, atomic, bombing, south\\
Topic 2 & germany, britain, france, europe, soviet\\ 
Topic 3 & court, justice, commission, law, attorney\\ 
Topic 4 & tax, oil, prices, petroleum, industry\\ 
Topic 5 & coal, union, strike, workers, miners\\ \bottomrule
\end{tabular}
\caption{Randomly sampled topics and top keywords derived from a 50-topic LDA model trained on 28K COHA articles identified as political using \textsc{Adaptive Ensembling}.} 
\label{tab:ae-topics}
\end{table}

\begin{figure}[t]
    \centering
    \includegraphics[scale=0.55]{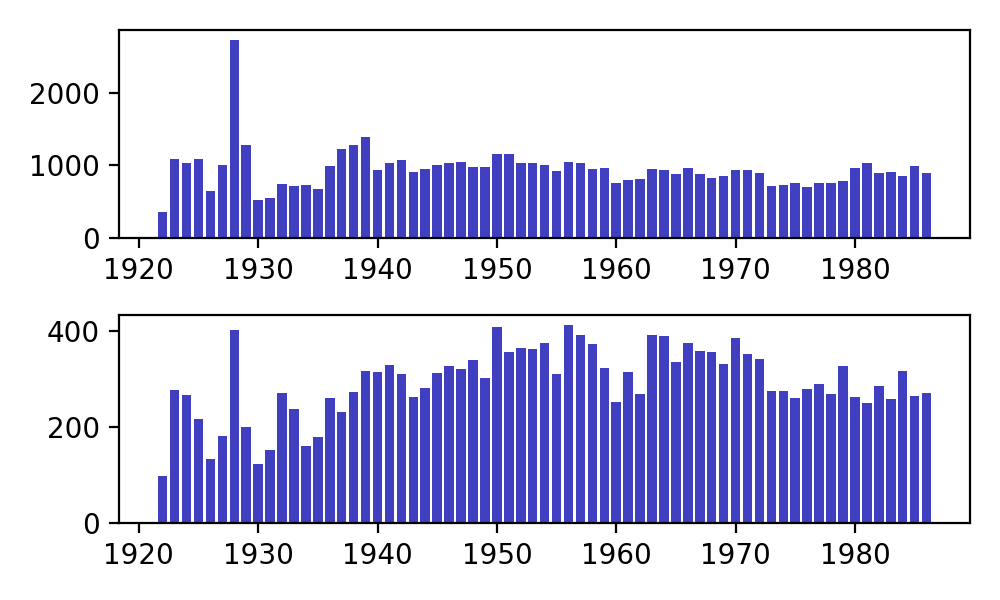}
    \caption{Document counts per decade for the original COHA corpus (top) and the extracted political subcorpus (bottom).}
    \label{fig:year}
\end{figure}

\begin{figure}[t]
    \centering
    \includegraphics[scale=0.37]{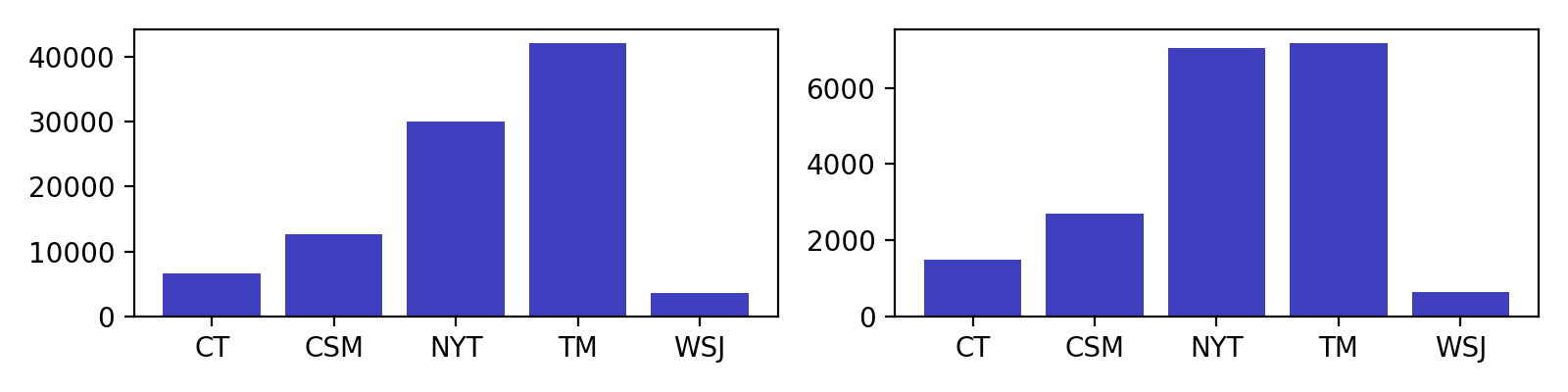}
    \caption{Document counts per news source for the original COHA corpus (left) and the extracted political subcorpus (right). The sources displayed include Chicago Tribute (CT), Christian Science Monitor (CSM), New York Times (NYT), Time Magazine (TM), and Wall Street Journal (WSJ).}
    \label{fig:pub}
\end{figure}

\paragraph{Does adaptive ensembling preserve the integrity of the original corpus?}
In order for political scientists to effectively study latent variables---such as political polarization---over time, the extracted political subcorpus must contain a similar integrity as the original corpus. That is, the subcorpus' distribution of documents across years and sources must relatively match that of the original corpus. First, we analyze the document counts for each decade bin, shown in Figure \ref{fig:year}. The political subcorpus shows a relatively consistent count across the deacdes, notably also capturing salient peaks from the 1920-1930s. Next, we analyze the document counts for each news source. Once again, the political subcorpus features documents from \textit{all} sources that appear in the original corpus. In addition, the varied distribution across sources is also captured; Time Magazine (TM) has the most documents whereas Wall Street Journal (WSJ) has the least documents. Together, these results show that the resulting subcorpus is amenable for political science research as it exhibits important characteristics derived from the original COHA corpus.

\section{Related Work}

Early approaches for unsupervised domain adaptation use shared autoencoders to create cross-domain representations \cite{glorot2011domain,chen2012marginalized}. More recently, \citet{ganin2016domain} introduce a new paradigm that create domain-invariant representations through adversarial training. 
This has gained popularity in NLP~\cite{zhang2017aspect,fu2017domain,chen2018adversarial}, however the difficulties of adversarial training are well-established \cite{salimans2016improved,arjovsky2017towards}. Consistency regularization methods (e.g., self-ensembling) outperform adversarial methods on visual semi-supervised and domain adaptation tasks \cite{athiwaratkun2018there}, but have rarely been applied to textual data \cite{ko2018domain}. Finally, \citet{huang2018examining} establish the feasibility of using domain adaptation to label documents from discrete time periods. Our work departs from previous work by proposing an adaptive, time-aware approach to consistency regularization provisioned with causal convolutional networks.

\section{Conclusion}

We present {\em adaptive ensembling}, an unsupervised domain adaptation framework capable of identifying political texts for a multi-source, diachronic corpus by only leveraging supervision from a single-source, modern corpus. Our methods outperform strong benchmarks on both binary and multi-label classification tasks. We release our system, as well as an expert-annotated set of political articles from COHA, to facilitate domain adaptation research in NLP and political science research on public opinion over time.

\section*{Acknowledgments}

The authors acknowledge the Texas Advanced Computing Center (TACC) at The University of Texas at Austin for providing HPC resources used to conduct this research. Thanks as well to Greg Durrett, Katrin Erk, and the anonymous reviewers for their helpful comments. This work was partially supported by the NSF Grant IIS-1850153.

\bibliography{references}

\begin{thebibliography}{43}
\expandafter\ifx\csname natexlab\endcsname\relax\def\natexlab#1{#1}\fi

\bibitem[{Acree et~al.(2018)Acree, Gross, Smith, Sim, and
  Boydstun}]{acree2018etch}
Brice~DL Acree, Justin~H Gross, Noah~A Smith, Yanchuan Sim, and Amber~E
  Boydstun. 2018.
\newblock Etch-a-sketching: Evaluating the post-primary rhetorical moderation
  hypothesis.
\newblock \emph{American Politics Research}.

\bibitem[{Arjovsky and Bottou(2017)}]{arjovsky2017towards}
Martin Arjovsky and L{\'e}on Bottou. 2017.
\newblock Towards principled methods for training generative adversarial
  networks.
\newblock In \emph{International Conference on Learning Representations}.

\bibitem[{Athiwaratkun et~al.(2019)Athiwaratkun, Finzi, Izmailov, and
  Wilson}]{athiwaratkun2018there}
Ben Athiwaratkun, Marc Finzi, Pavel Izmailov, and Andrew~Gordon Wilson. 2019.
\newblock There are many consistent explanations of unlabeled data: Why you
  should average.
\newblock In \emph{International Conference on Learning Representations}.

\bibitem[{Bai et~al.(2018)Bai, Kolter, and Koltun}]{bai2018empirical}
Shaojie Bai, J~Zico Kolter, and Vladlen Koltun. 2018.
\newblock An empirical evaluation of generic convolutional and recurrent
  networks for sequence modeling.
\newblock \emph{arXiv preprint arXiv:1803.01271}.

\bibitem[{Bai et~al.(2019)Bai, Kolter, and Koltun}]{bai2018trellis}
Shaojie Bai, J.~Zico Kolter, and Vladlen Koltun. 2019.
\newblock Trellis networks for sequence modeling.
\newblock In \emph{International Conference on Learning Representations}.

\bibitem[{Baldassarri and Gelman(2008)}]{baldassarri2008partisans}
Delia Baldassarri and Andrew Gelman. 2008.
\newblock Partisans without constraint: Political polarization and trends in
  american public opinion.
\newblock \emph{American Journal of Sociology}, 114(2):408--446.

\bibitem[{Bau et~al.(2019)Bau, Belinkov, Sajjad, Durrani, Dalvi, and
  Glass}]{bau2018identifying}
Anthony Bau, Yonatan Belinkov, Hassan Sajjad, Nadir Durrani, Fahim Dalvi, and
  James Glass. 2019.
\newblock Identifying and controlling important neurons in neural machine
  translation.
\newblock In \emph{International Conference on Learning Representations}.

\bibitem[{Baum and Potter(2008)}]{baum2008relationships}
Matthew~A Baum and Philip~BK Potter. 2008.
\newblock The relationships between mass media, public opinion, and foreign
  policy: Toward a theoretical synthesis.
\newblock \emph{Annual Review of Political Science}, 11:39--65.

\bibitem[{Bengio et~al.(2009)Bengio, Louradour, Collobert, and
  Weston}]{bengio2009curriculum}
Yoshua Bengio, J{\'e}r{\^o}me Louradour, Ronan Collobert, and Jason Weston.
  2009.
\newblock Curriculum learning.
\newblock In \emph{Proceedings of the 26th Annual International Conference on
  Machine Learning}, pages 41--48.

\bibitem[{Bishop(2004)}]{bishop2004illusion}
George~F Bishop. 2004.
\newblock \emph{The illusion of public opinion: Fact and artifact in American
  public opinion polls}.
\newblock Rowman \& Littlefield Publishers.

\bibitem[{Blei et~al.(2003)Blei, Ng, and Jordan}]{blei2003latent}
David~M Blei, Andrew~Y Ng, and Michael~I Jordan. 2003.
\newblock Latent dirichlet allocation.
\newblock \emph{Journal of Machine Learning Research}, 3(Jan):993--1022.

\bibitem[{Campbell et~al.(1980)Campbell, Converse, Miller, and
  Stokes}]{campbell1980american}
Angus Campbell, Philip~E Converse, Warren~E Miller, and Donald~E Stokes. 1980.
\newblock \emph{The {A}merican Voter}.
\newblock University of Chicago Press.

\bibitem[{Chen et~al.(2012)Chen, Xu, Weinberger, and
  Sha}]{chen2012marginalized}
Minmin Chen, Zhixiang Xu, Kilian~Q. Weinberger, and Fei Sha. 2012.
\newblock Marginalized denoising autoencoders for domain adaptation.
\newblock In \emph{Proceedings of the 29th International Conference on
  International Conference on Machine Learning}, pages 1627--1634.

\bibitem[{Chen et~al.(2018)Chen, Sun, Athiwaratkun, Cardie, and
  Weinberger}]{chen2018adversarial}
Xilun Chen, Yu~Sun, Ben Athiwaratkun, Claire Cardie, and Kilian Weinberger.
  2018.
\newblock Adversarial deep averaging networks for cross-lingual sentiment
  classification.
\newblock \emph{Transactions of the Association for Computational Linguistics},
  6:557--570.

\bibitem[{Cohen(1960)}]{cohen1960coefficient}
Jacob Cohen. 1960.
\newblock A coefficient of agreement for nominal scales.
\newblock \emph{Educational and psychological measurement}, 20(1):37--46.

\bibitem[{Davidov et~al.(2014)Davidov, Meuleman, Cieciuch, Schmidt, and
  Billiet}]{davidov2014measurement}
Eldad Davidov, Bart Meuleman, Jan Cieciuch, Peter Schmidt, and Jaak Billiet.
  2014.
\newblock Measurement equivalence in cross-national research.
\newblock \emph{Annual review of sociology}, 40:55--75.

\bibitem[{Davies(2008)}]{davies2008corpus}
Mark Davies. 2008.
\newblock The corpus of contemporary american english: 450 million words,
  1990-present.

\bibitem[{Elkins and Shaffer(2019)}]{elkins2019evaluation}
Zachary Elkins and Robert Shaffer. 2019.
\newblock On measuring textual similarity.
\newblock Work in progress.

\bibitem[{French et~al.(2018)French, Mackiewicz, and Fisher}]{french2017self}
Geoff French, Michal Mackiewicz, and Mark Fisher. 2018.
\newblock Self-ensembling for visual domain adaptation.
\newblock In \emph{International Conference on Learning Representations}.

\bibitem[{Fu et~al.(2017)Fu, Nguyen, Min, and Grishman}]{fu2017domain}
Lisheng Fu, Thien~Huu Nguyen, Bonan Min, and Ralph Grishman. 2017.
\newblock Domain adaptation for relation extraction with domain adversarial
  neural network.
\newblock In \emph{Proceedings of the Eighth International Joint Conference on
  Natural Language Processing (Volume 2: Short Papers)}, pages 425--429.

\bibitem[{Ganin et~al.(2016)Ganin, Ustinova, Ajakan, Germain, Larochelle,
  Laviolette, Marchand, and Lempitsky}]{ganin2016domain}
Yaroslav Ganin, Evgeniya Ustinova, Hana Ajakan, Pascal Germain, Hugo
  Larochelle, Fran{\c{c}}ois Laviolette, Mario Marchand, and Victor Lempitsky.
  2016.
\newblock Domain-adversarial training of neural networks.
\newblock \emph{The Journal of Machine Learning Research}, 17(1):2096--2030.

\bibitem[{Glorot et~al.(2011)Glorot, Bordes, and Bengio}]{glorot2011domain}
Xavier Glorot, Antoine Bordes, and Yoshua Bengio. 2011.
\newblock Domain adaptation for large-scale sentiment classification: A deep
  learning approach.
\newblock In \emph{Proceedings of the 28th International Conference on Machine
  Learning}, pages 513--520.

\bibitem[{Goodin(2009)}]{goodin2009oxford}
Robert~E Goodin. 2009.
\newblock \emph{The Oxford handbook of political science}, volume~11.
\newblock Oxford University Press.

\bibitem[{Hochreiter and Schmidhuber(1997)}]{hochreiter1997long}
Sepp Hochreiter and J{\"u}rgen Schmidhuber. 1997.
\newblock Long short-term memory.
\newblock \emph{Neural computation}, 9(8):1735--1780.

\bibitem[{Huang and Paul(2018)}]{huang2018examining}
Xiaolei Huang and Michael~J Paul. 2018.
\newblock Examining temporality in document classification.
\newblock In \emph{Proceedings of the 56th Annual Meeting of the Association
  for Computational Linguistics (Volume 2: Short Papers)}, pages 694--699.

\bibitem[{Kalchbrenner et~al.(2016)Kalchbrenner, Espeholt, Simonyan, Oord,
  Graves, and Kavukcuoglu}]{kalchbrenner2016neural}
Nal Kalchbrenner, Lasse Espeholt, Karen Simonyan, Aaron van~den Oord, Alex
  Graves, and Koray Kavukcuoglu. 2016.
\newblock Neural machine translation in linear time.
\newblock \emph{arXiv preprint arXiv:1610.10099}.

\bibitem[{Kim(2014)}]{kim2014convolutional}
Yoon Kim. 2014.
\newblock Convolutional neural networks for sentence classification.
\newblock In \emph{Proceedings of the 2014 Conference on Empirical Methods in
  Natural Language Processing}, pages 1746--1751.

\bibitem[{Kingma and Ba(2015)}]{kingma2014adam}
Diederik~P Kingma and Jimmy Ba. 2015.
\newblock Adam: A method for stochastic optimization.
\newblock In \emph{International Conference for Learning Representations}.

\bibitem[{Ko et~al.(2019)Ko, Durrett, and Li}]{ko2018domain}
Wei-Jen Ko, Greg Durrett, and Junyi~Jessy Li. 2019.
\newblock Domain agnostic real-valued specificity prediction.
\newblock In \emph{Proceedings of the AAAI Conference on Artificial
  Intelligence}.

\bibitem[{Laine and Aila(2017)}]{laine2016temporal}
Samuli Laine and Timo Aila. 2017.
\newblock Temporal ensembling for semi-supervised learning.
\newblock In \emph{International Conference for Learning Representations}.

\bibitem[{Long et~al.(2015)Long, Shelhamer, and Darrell}]{long2015fully}
Jonathan Long, Evan Shelhamer, and Trevor Darrell. 2015.
\newblock Fully convolutional networks for semantic segmentation.
\newblock In \emph{Proceedings of the IEEE Conference on Computer Vision and
  Pattern Recognition}, pages 3431--3440.

\bibitem[{de~Marchi et~al.(2018)de~Marchi, Dorsey, and Ensley}]{de2018policy}
Scott de~Marchi, Spencer Dorsey, and Michael~J Ensley. 2018.
\newblock Policy and the structure of roll call voting in the us house.
\newblock \emph{Available at SSRN 3262316}.

\bibitem[{McCombs(2018)}]{mccombs2018setting}
Maxwell McCombs. 2018.
\newblock \emph{Setting the agenda: Mass media and public opinion}.
\newblock John Wiley \& Sons.

\bibitem[{Pennington et~al.(2014)Pennington, Socher, and
  Manning}]{pennington2014glove}
Jeffrey Pennington, Richard Socher, and Christopher Manning. 2014.
\newblock Glove: Global vectors for word representation.
\newblock In \emph{Proceedings of the 2014 Conference on Empirical Methods in
  Natural Language Processing}, pages 1532--1543.

\bibitem[{Salimans et~al.(2016)Salimans, Goodfellow, Zaremba, Cheung, Radford,
  and Chen}]{salimans2016improved}
Tim Salimans, Ian Goodfellow, Wojciech Zaremba, Vicki Cheung, Alec Radford, and
  Xi~Chen. 2016.
\newblock Improved techniques for training {GAN}s.
\newblock In \emph{Advances in Neural Information Processing Systems 29}, pages
  2234--2242.

\bibitem[{Sandhaus(2008)}]{sandhaus2008new}
Evan Sandhaus. 2008.
\newblock The {New York Times} annotated corpus.
\newblock \emph{Linguistic Data Consortium, Philadelphia}, 6(12):e26752.

\bibitem[{Srivastava et~al.(2014)Srivastava, Hinton, Krizhevsky, Sutskever, and
  Salakhutdinov}]{srivastava2014dropout}
Nitish Srivastava, Geoffrey Hinton, Alex Krizhevsky, Ilya Sutskever, and Ruslan
  Salakhutdinov. 2014.
\newblock Dropout: a simple way to prevent neural networks from overfitting.
\newblock \emph{The Journal of Machine Learning Research}, 15(1):1929--1958.

\bibitem[{Tarvainen and Valpola(2017)}]{tarvainen2017mean}
Antti Tarvainen and Harri Valpola. 2017.
\newblock Mean teachers are better role models: Weight-averaged consistency
  targets improve semi-supervised deep learning results.
\newblock In \emph{Advances in Neural Information Processing Systems 30}, pages
  1195--1204.

\bibitem[{Vaswani et~al.(2017)Vaswani, Shazeer, Parmar, Uszkoreit, Jones,
  Gomez, Kaiser, and Polosukhin}]{vaswani2017attention}
Ashish Vaswani, Noam Shazeer, Niki Parmar, Jakob Uszkoreit, Llion Jones,
  Aidan~N Gomez, {\L}ukasz Kaiser, and Illia Polosukhin. 2017.
\newblock Attention is all you need.
\newblock In \emph{Advances in Neural Information Processing Systems 30}, pages
  5998--6008.

\bibitem[{Yang et~al.(2017)Yang, Hu, Salakhutdinov, and
  Berg-Kirkpatrick}]{yang2017improved}
Zichao Yang, Zhiting Hu, Ruslan Salakhutdinov, and Taylor Berg-Kirkpatrick.
  2017.
\newblock Improved variational autoencoders for text modeling using dilated
  convolutions.
\newblock In \emph{Proceedings of the 34th International Conference on Machine
  Learning}, pages 3881--3890.

\bibitem[{Yang et~al.(2016)Yang, Yang, Dyer, He, Smola, and
  Hovy}]{yang2016hierarchical}
Zichao Yang, Diyi Yang, Chris Dyer, Xiaodong He, Alex Smola, and Eduard Hovy.
  2016.
\newblock Hierarchical attention networks for document classification.
\newblock In \emph{Proceedings of the 2016 Conference of the North American
  Chapter of the Association for Computational Linguistics: Human Language
  Technologies}, pages 1480--1489.

\bibitem[{Zaller et~al.(1992)}]{zaller1992nature}
John~R Zaller et~al. 1992.
\newblock \emph{The nature and origins of mass opinion}.
\newblock Cambridge University Press.

\bibitem[{Zhang et~al.(2017)Zhang, Barzilay, and Jaakkola}]{zhang2017aspect}
Yuan Zhang, Regina Barzilay, and Tommi Jaakkola. 2017.
\newblock Aspect-augmented adversarial networks for domain adaptation.
\newblock \emph{Transactions of the Association for Computational Linguistics},
  5:515--528.

\end{thebibliography}
\bibliographystyle{acl_natbib}

\clearpage
\appendix

\section{LDA Topic Model}
\label{appendix:topic}

We experimented with a range of hyperparameters to ensure the Latent Dirichlet Allocation (LDA) model was best optimized for our datasets, leveraging the Gensim\footnote{https://radimrehurek.com/gensim/} library. In particular, we removed all stopwords, extremely rare words (tail 10-20\% from a unigram distribution), and set the number of topics to 50.

\section{Self-Ensembling}
\label{sec:se-info}

The core intuition behind consistency regularization is that ensembled predictions are more likely to be correct than single predictions \cite{laine2016temporal,tarvainen2017mean}. To this end, \citet{laine2016temporal} introduce a \textbf{student} and \textbf{teacher} network that yield single predictions and ensembled predictions, respectively. 

After learning from labeled samples, the student may produce varying, dissimilar predictions for unlabeled samples due to the stochastic nature of optimization. One potential solution is to ensemble predictions across time to converge at the \textit{most likely} prediction \cite{laine2016temporal}. \citet{tarvainen2017mean} improve upon this method by showing that ensembling parameters (as opposed to predictions) results in better predictions. Because the teacher's parameters are smoothed with the student's learned parameters at each iteration, the teacher effectively becomes an ensemble of the student across time.

Further, to ensure that the features learned from the labeled samples are compatible with the unlabeled samples, \citet{laine2016temporal,tarvainen2017mean,french2017self} motivate a consistency-enforcing approach to bring the student and teacher's predictions closer together. In essence, if a feature learned from samples in the labeled domain is incompatible with samples in the unlabeled domain, the consistency (unsupervised) loss penalizes its incompatibility. Therefore, the interplay between these two networks creates a robust, domain-invariant feature space that characterizes both labeled and unlabeled samples \cite{french2017self}. A detailed visualization of the training procedure is presented in Figure \ref{fig:fw-img} in the main body of this paper.

\section{NYT Descriptors}
\label{sec:nyt-data}

We build a list of ``political'' descriptors in NYT to determine (a) which labels we can or cannot sample non-political documents from; and (b) which descriptors fall under the three areas of political science we consider for our multi-label task (American Government, Political Economy, and International Relations).

Because documents can be tagged with multiple descriptors, we build a list of descriptors whose documents have significant overlap with \textsc{US Politics \& Government}. The second author, a political science graduate student,  filtered this list to 57 descriptors that are political in nature.

For (a), we sample 4,600 non-political documents whose descriptors do not overlap with the 57 political descriptors described above. For (b), the same political science graduate student assigns each descriptor with one or more area labels. We use this label information to build an NYT dataset for our tasks. The 57 political descriptors and their corresponding area labels are tabulated in Table \ref{tab:data}.

\section{Expert-Annotated Dataset}
\label{sec:expa-data}

\begin{table}[t]
\centering
\small
\begin{tabular}{c|ccc|c}
\toprule
 & \multicolumn{3}{c|}{Political} & Non-Political \\ \midrule
 & AG & PE & IR &  \\ \midrule
Train & 333 & 8 & 156 & 497 \\ 
Dev & 82 & 1 & 33 & 116 \\ 
Test & 125 & 8 & 47 & 208 \\ \bottomrule
\end{tabular}
\caption{Distribution of train (In-Domain benchmark \textit{only}), dev, test documents in our expert-annotated COHA subcorpus. For political documents, we break down the distribution into American Government (AG), Political Economy (PE), and International Relations (IR).}
\label{tab:coha}
\end{table}

To create an initial COHA subcorpus of 56,000 documents (8,000 per decade), we sample from the following news sources that consistently appear in across decades: Chicago Tribune, Christian Science Monitor, New York Times, Time Magazine, and Wall Street Journal. Note that these NYT articles (up to year 1986) do not appear in the NYT annotated corpus~\cite{sandhaus2008new} (starting from year 1987), which we used as our source, training dataset.

From this subcorpus, we perform additional steps to create an expert-annotated dataset (\S\ref{sec:data}). Label distributions for our dataset are presented in Table \ref{tab:coha}. Although political economy (PE) is severely underrepresented, we experimentally find that these documents have salient features and are not as difficult to classify. In addition, we employ class imbalance penalties to prevent our model from ignoring these documents.

The source dataset (NYT) was already annotated; to ensure label agreement with our target dataset (COHA), we sampled documents from the source dataset and had our political science graduate students label them to compare against the original label. There were minimal problems here---because NYT has fine-grained labels for their documents, the politically-labeled articles were clearly political and vice-versa.

The target datatset (COHA) was divided into halves and each political science graduate student annotated a half. Prior to annotation, they agreed upon a set of rules to minimize bias in the annotation process. In addition, both of them worked side-by-side during all annotation periods, so they were able to ask each other's opinion in case there was confusion. We also took measures to ensure label correctness after annotation was completed. Each political science graduate student sampled a batch of their political and non-political annotations and sent it to the other to evaluate. Again, there was not much disagreement here as the rules decided upon in the beginning were sufficient to cover most edge cases. Quantitatively, Cohen's $\kappa = 0.95$ as calculated on a mutually annotated subset \cite{cohen1960coefficient}.

\begin{table}[]
\centering
\small
\begin{tabular}{p{5cm}|c|c|c}
 & \multicolumn{3}{c}{{\bf Area Label}} \\ \hline
{\bf Topic} & {\bf AG} & {\bf PE} & {\bf IR} \\ \hline
International Relations &  &  & \checkmark \\ \hline
Presidents and Presidency (US) & \checkmark &  &  \\ \hline
Presidential Elections (US) & \checkmark &  &  \\ \hline
War and Revolution &  &  & \checkmark \\ \hline
Presidential Election of 2000 & \checkmark &  &  \\ \hline
Presidential Election of 2004 & \checkmark &  &  \\ \hline
Law and Legislation & \checkmark &  &  \\ \hline
Civil War and Guerrilla Warfare &  &  & \checkmark \\ \hline
International Trade and World Market &  & \checkmark &  \\ \hline
Presidential Election of 1996 & \checkmark &  &  \\ \hline
Public Opinion &  &  &  \\ \hline
Economic Conditions and Trends &  & \checkmark &  \\ \hline
Bombs and Explosives &  &  & \checkmark \\ \hline
Arms Sales Abroad &  &  & \checkmark \\ \hline
United States Economy &  & \checkmark &  \\ \hline
Missiles and Missile Defense Systems &  &  & \checkmark \\ \hline
Oil (Petroleum) and Gasoline &  & \checkmark &  \\ \hline
Appointments and Executive Changes & \checkmark &  &  \\ \hline
Foreign Service &  &  & \checkmark \\ \hline
Prisoners of War &  &  & \checkmark \\ \hline
War Crimes, Genocide and Crimes Against Humanity &  &  & \checkmark \\ \hline
Vice Presidents and Vice Presidency (US) & \checkmark &  &  \\ \hline
Arms Control and Limitation and Disarmament &  &  & \checkmark \\ \hline
Military Bases and Installations &  &  & \checkmark \\ \hline
Presidential Election of 2008 & \checkmark &  &  \\ \hline
Whitewater Case & \checkmark &  &  \\ \hline
Vietnam War & \checkmark &  & \checkmark \\ \hline
Governors (US) & \checkmark &  &  \\ \hline
Energy and Power &  & \checkmark &  \\ \hline
Stocks and Bonds &  & \checkmark &  \\ \hline
State of the Union Message (US) & \checkmark &  &  \\ \hline
Wages and Salaries &  & \checkmark &  \\ \hline
Church-State Relations & \checkmark &  &  \\ \hline
Shiite Muslims &  &  & \checkmark \\ \hline
Special Prosecutors (Independent Counsel) & \checkmark &  &  \\ \hline
White House (Washington, DC) & \checkmark &  &  \\ \hline
Federal Taxes (US) &  & \checkmark &  \\ \hline
Illegal Aliens & \checkmark &  &  \\ \hline
Social Security (US) & \checkmark &  &  \\ \hline
Political Prisoners & \checkmark &  & \checkmark \\ \hline
Watergate Affair & \checkmark &  &  \\ \hline
Government Employees & \checkmark &  &  \\ \hline
Sunni Muslims &  &  & \checkmark \\ \hline
Third World and Developing Countries &  &  & \checkmark \\ \hline
Customs (Tariff) &  & \checkmark &  \\ \hline
Welfare (US) &  & \checkmark &  \\ \hline
Gun Control & \checkmark &  &  \\ \hline
Global Warming & \checkmark &  &  \\ \hline
Interest Rates &  & \checkmark &  \\ \hline
Vetoes (US) & \checkmark &  &  \\ \hline
Futures and Options Trading &  & \checkmark &  \\ \hline
Attorneys General & \checkmark &  &  \\ \hline
Layoffs and Job Reductions &  & \checkmark &  \\ \hline
Nazi Policies Toward Jews and Minorities &  &  & \checkmark \\ \hline
Government Bonds &  & \checkmark &  \\ \hline
Police Brutality and Misconduct & \checkmark &  &  \\ \hline
Executive Privilege, Doctrine of & \checkmark &  &  \\ \hline
\end{tabular}
\caption{Political descriptors in NYT. Each descriptor is categorized under one or more political science areas: American Government (AG), Political Economy (PE), and International Relations (IR).}
\label{tab:data}
\end{table}

\end{document}